\documentclass{article}

\usepackage{graphicx}
\usepackage{comment}
\usepackage{amsmath,amssymb} 
\usepackage{multirow}
\usepackage{multicol}
\usepackage{subfigure}
\usepackage{algorithm}
\usepackage{algpseudocode}
\usepackage{bm}
\usepackage{booktabs}
\usepackage{microtype}
\usepackage[final]{corl_2020} 

\title{Reconfigurable Voxels: A New Representation for LiDAR-Based Point Clouds}

\author{
  Tai Wang, Xinge Zhu, Dahua Lin \vspace{0.5ex}\\
  The Chinese University of Hong Kong \vspace{0.5ex}\\
  \texttt{\{wt019, zx018, dhlin\}@ie.cuhk.edu.hk} \\
}

\begin{document}
\maketitle


\begin{abstract}

LiDAR is an important method for autonomous driving systems to sense the environment. 
The point clouds obtained by LiDAR typically exhibit sparse and irregular distribution, 
thus posing great challenges to the detection of 3D objects, especially those that 
are small and distant.
To tackle this difficulty, we propose Reconfigurable Voxels, 
a new approach to constructing representations from 3D point clouds.
Specifically, we devise a biased random walk scheme, which adaptively covers each neighborhood
with a fixed number of voxels based on the local spatial distribution
and produces a representation by integrating the points 
in the chosen neighbors.
We found empirically that this approach effectively improves the stability of voxel 
features, especially for sparse regions. 
Experimental results on multiple benchmarks, including nuScenes, Lyft, and KITTI, show that 
this new representation can remarkably improve the detection performance for small and 
distant objects, without incurring noticeable overhead costs.
\end{abstract}
\vspace{-2ex}



\section{Introduction}
\label{sec:introduction}

LiDAR has been widely used in driver assistance or autonomous driving systems~\cite{LIDAR,AUV}, 
which senses the environment via reflected laser light and produces 3D point clouds 
as the output. 
Compared to conventional 3D data, 
\emph{e.g.} those obtained by 3D scanner for object modeling~\cite{3DShapeNets,shapenet2015}, 
the 3D point clouds derived by LiDAR are usually much more sparse and irregular. 
Therefore, effective handling of such data requires new methods -- in particular
new representations tailored to LiDAR's special characteristics.


Existing approaches to 3D point cloud representation mainly follow two streams:
\emph{point-based} and \emph{voxel-based}. 
Point-based methods~\cite{PointNet,PointCNN,RSNet,Interp-Conv,PointRCNN}, among which PointNet~\cite{PointNet} is a representative, 
focus on the processing of individual points and integrate the information on top.
Due to the narrow focus in the initial processing stage, point-based methods often 
lack the capability of capturing large spatial structures.
Voxel-based methods~\cite{3DShapeNets,OctNet,VoxelNet,ssn}, instead, begin with the space. 
Specifically, they quantize a 3D space into cells and process the information 
based on the cells instead of individual points. 
While this allows spatial distributions of greater scale to be captured, 
the tradeoff between representation precision and computational complexity 
remains an open problem.
This problem is especially crucial for sparsely distributed point clouds. 

\begin{figure}
   \centering
   \includegraphics[width=1.0\linewidth]{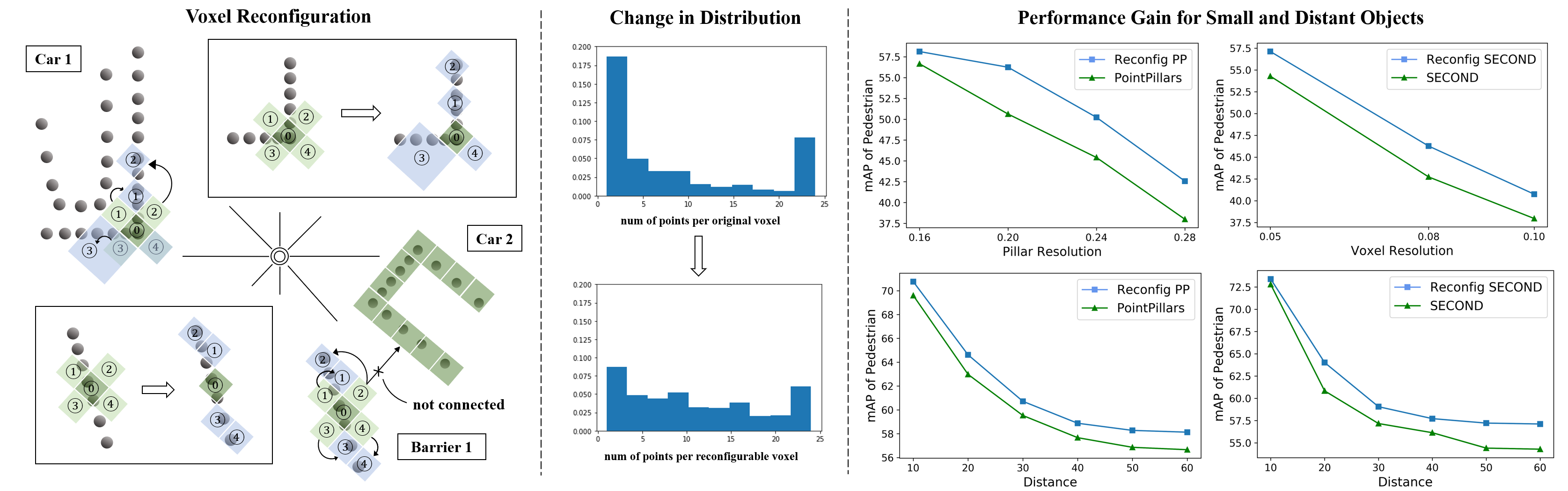}
   \vspace{-0.5cm}
   \caption{\textbf{Left:} we take the original voxel together with its 4 neighbor voxels as a whole to encode features. Light green and blue voxels represent neighbors before and after reconfiguration respectively. Note that the transition of neighbors is only carried out on the same connected component composed of non-empty voxels. Without loss of generality, reconfiguration here operates on X-Y plane. Extension to 3D is straightforward. \textbf{Middle:} The reconfigurable voxels greatly improve the imbalance of sampling points in different voxels, thus encoding more robust features in sparse regions. \textbf{Right:} Our method can consistently improve the detection performance for small and distant objects under multiple settings and frameworks on KITTI.}
   \label{fig: teaser}
   \vspace{-2.8ex}
\end{figure}

In this work, we choose to follow the voxel-based approach, due to its inherent strength 
in modeling spatial distributions, while aiming to tackle the difficulties caused by 
the sparsity and irregularity in LiDAR data. 
Specifically, we propose \emph{Reconfigurable Voxels}, a generic voxel-based 
representation. 
As shown in Fig.~\ref{fig: teaser}, for each voxel, 
it adaptively \emph{reconfigures} its neighborhood
through a biased random walk so as to cover its surrounding regions more 
effectively, and then derives an embedding thereon.

The proposed method has several appealing properties:
(1) \emph{More stable representation.}
By constructing features upon an adaptive neighborhood, it effectively mitigates 
the difficulties caused by sparsity and irregularity, \emph{e.g.} voxels with few or even 
no points, thus resulting in more stable features. 
(2) \emph{Strong locality.}
While allowed to be stretched, the reconfigured neighborhood remains within 
a surrounding region of the target location, and therefore still preserves 
strong locality. This is important for capturing local structures. 
(3) \emph{High efficiency.}
It is noteworthy that the construction of the voxel neighborhoods can be done 
in one traversal of the dataset and then fixed. Compared to the overall 
computing cost, this additional overhead is insignificant.


We evaluated the proposed representation method on multiple benchmarks of 
3D detection, including nuScenes~\cite{nuScenes}, Lyft~\cite{Lyft} and KITTI~\cite{KITTI}. 
On these datasets, it consistently achieved significant performance gains.
Moreover, our study also shows that \emph{Reconfigurable Voxels} can effectively 
handle the sparse point clouds, thus substantially improving the 
capability of detecting small and distant objects: it can boost the performance of most small objects by over 2\% mAP on all datasets and objects over 20 meters away by 4.4\% NDS on nuScenes. 


\vspace{-1ex}
\section{Related Work}
\label{sec:related}
\noindent\textbf{3D Object Detection}~~ The problem of 3D object detection has been widely explored before deep learning approaches emerged. 
Firstly, work focusing on indoor scenes includes: \cite{holistic} modeled contextual relationship to guide object detection; \cite{sliding-shapes} designed sliding-shapes to realize detection in RGB-D images; and VoteNet~\cite{VoteNet} utilized a reformulation of Hough voting in 3D case, \emph{etc.}

Among work for autonomous vehicles, although image-based methods~\cite{FQNet,PseudoLiDAR,StereoRCNN} have made great progress, their performance is still far behind LiDAR-based methods. The methods using LiDAR data can be divided into two categories according to the data types used: the methods using multimodal data and the methods using only LiDAR data. The first batch of methods~\cite{MV3D,AVOD} resolved this problem by fusing features extracted from images and projections of point clouds, which reduced 3D problems to 2D cases. 
Then with PointNet~\cite{PointNet} proposed, it became possible to extract features directly from point cloud data. Earlier works deploying this backbone like \cite{F-PointNet} used 2D detection guided frustum to reduce search space. The other methods follow two streams: voxel-based and point-based. Among voxel-based methods, \cite{PIXOR,ComplexYOLO} utilized hand-crafted features to detect objects in bird view map, while VoxelNet and SECOND~\cite{VoxelNet,SECOND} directly processed 3D partitioned voxels, used PointNet to encode, and trained them as a module in the end-to-end framework. PointPillars~\cite{PointPillars} simplified the representation to pillar, thus obtaining a bird view pseudo image after encoding, and further improved the efficiency with 2D convolution. 
Point-based methods~\cite{PointRCNN,STD,FastPointRCNN}, instead, designed frameworks to extract proposals and detected objects in point level based on scene segmentation module, but the number of points needed to process is always a limitation to these methods. 
Therefore, our work carries on the exploration in voxel-based methods.
Although recent work~\cite{PVRCNN,DynamicVoxel,HVNet} began to explore to encode features more effectively from better representation, it is not divorced from the original voxel layout or simply fuses multi-scale information. In comparison, our  reconfigurable voxels is a kind of deformable voxel representation which is constructed in reasonable local space according to the spatial distribution of points, so as to depict the shape of objects implicitly.

\noindent\textbf{Voxel-based Learning on Point Cloud}~~ Utilizing voxel as the basic representation is an intuitive way to migrate 2D methods to 3D problems.
To mention only a few, ~\cite{3DShapeNets} proposed 3D ShapeNets to achieve object recognition and shape completion. \cite{VoxNet} improved it with fewer input parameters. \cite{OctNet,O-CNN} used octree structure to improve the efficiency problem caused by 3D convolution.
Nevertheless, these works only focus on the case of a single object. Given the difference between the point cloud of CAD models and LiDAR-based data, how to transfer these ideas is still an open question.

\noindent\textbf{Deformable Convolutional Networks}~~ The traditional convolution can be regarded as a fixed kernel executing point-wise inner-product with the corresponding content of the image at a specific location. In the 2D case, deformation modeling is a common and principled problem, and there are many works like \cite{DeformConv} targeting it and designing variant convolutions to extract features flexibly. 
In addition, some other works considered to sample in the kernel space without changing the theoretical receptive fields, such as \cite{DeformKernel} in the 2D case and \cite{KPConv,DeformFilter} in 3D point clouds. In comparison, while our devised reconfiguration is similar with deformation, the motivation is not the same: the deformation and scale problems in 2D do not exist in 3D cases. The problem we try to tackle is the difficulty of detecting small and distant objects caused by the irregular spatial distribution of LiDAR-based point clouds. It is intuitively more straightforward to introduce deformation into the point-to-voxel process instead of modifying the convolution operation on the voxel feature maps.
\begin{figure*}
\begin{center}
\includegraphics[width=1.0\linewidth]{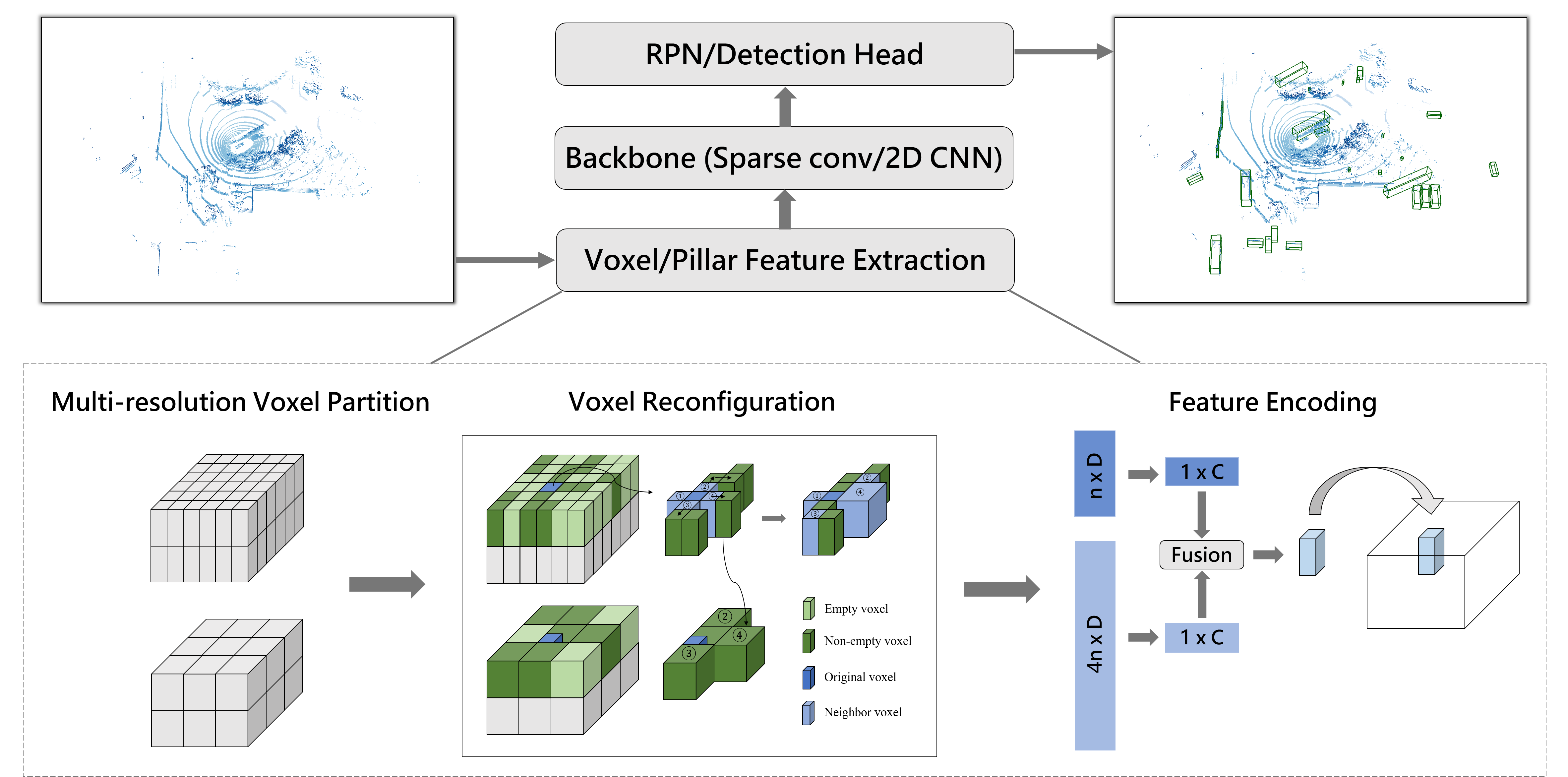}
\end{center}
\vspace{-1.5ex}
   \caption{An overview of our pipeline. Reconfigurable voxels is a generic representation which can be exploited when extracting voxel features. With feature fusion of the original voxel and its 4 reconfigured neighbors, we can adaptively encode local shape without modifying the data structure of the following operations. Here we show voxel reconfiguration module in the multi-resolution case. Arrows between voxels indicate transitions of neighbors. See more details in Sec.~\ref{multires}}
\label{fig: pipeline}
\vspace{-3ex}
\end{figure*}

\vspace{-1ex}
\section{Approach}
\noindent\textbf{Overview}~~ How to construct an efficient representation from sparsely and irregularly distributed point cloud is a key problem for scene understanding tasks, like 3D detection in autonomous driving. In general, a voxel-based 3D detection framework groups raw point cloud into voxels, applies voxel feature encoding layers and scatters them back for the subsequent convolutional backbone and prediction of 3D bounding boxes.
Our reconfigurable voxels is a generic representation when partitioning the space, which encodes local information more effectively by covering voxel neighbors based on the spatial distribution.
Next in this section, we will elaborate the construction method and technical details of reconfigurable voxels in turn, and finally extend the single-resolution case to multi-resolution, making the whole design more flexible and robust.

\subsection{Construction of Reconfigurable Voxels}
To address the problem caused by sparsity and irregularity, a simple idea is to allow the existence of voxels in different sizes, but this easily destroys the data structure of subsequent computations, and thus is not conducive to maintaining the real-time performance of the algorithm. Therefore, we propose that on the basis of primitive voxel partition, the original voxel can cover its surrounding regions more effectively by reconfiguring its neighborhood based on the local spatial distribution. 

Specifically, the process of constructing reconfigurable voxels is as follows: Firstly, the whole scene is divided into voxels of the same size, and the index of each neighbor is recorded in the process of partitioning.
Thus, the construction of graphs can be completed in a one-time traversal process. Subsequently, we make every neighbor of each voxel carry out a biased random walk. A mechanism is designed to make the neighbors walk to voxels with denser point clouds. Finally, we compose these reconfigured neighbors with the original voxel, extract features, fuse them, and scatter the final features back to the original location. See the process in Fig.~\ref{fig: pipeline}.

It can be seen that this voxel partition process does not affect the operations of the subsequent backbone. Meanwhile, through the reconfiguration of neighbors, the vulnerability of voxel features in sparse regions is improved. Note that this process is free of learning parameters, which thus avoids possible indifferentiable problems when how to carry out random walk between voxels needs to be learned and maintains our end-to-end training.
In addition, because these neighbors are only allowed to walk on the same connected component, it basically ensures that they will be in the adjacent area instead of freely running across the open area to other irrelevant objects, which leverages the sparsity of LiDAR data and depicts local shape of objects implicitly.

\subsection{Biased Random Walking Neighbors} \label{sec:biasedRW}
\begin{figure*}
    \centering
    \includegraphics[width=1.0\linewidth]{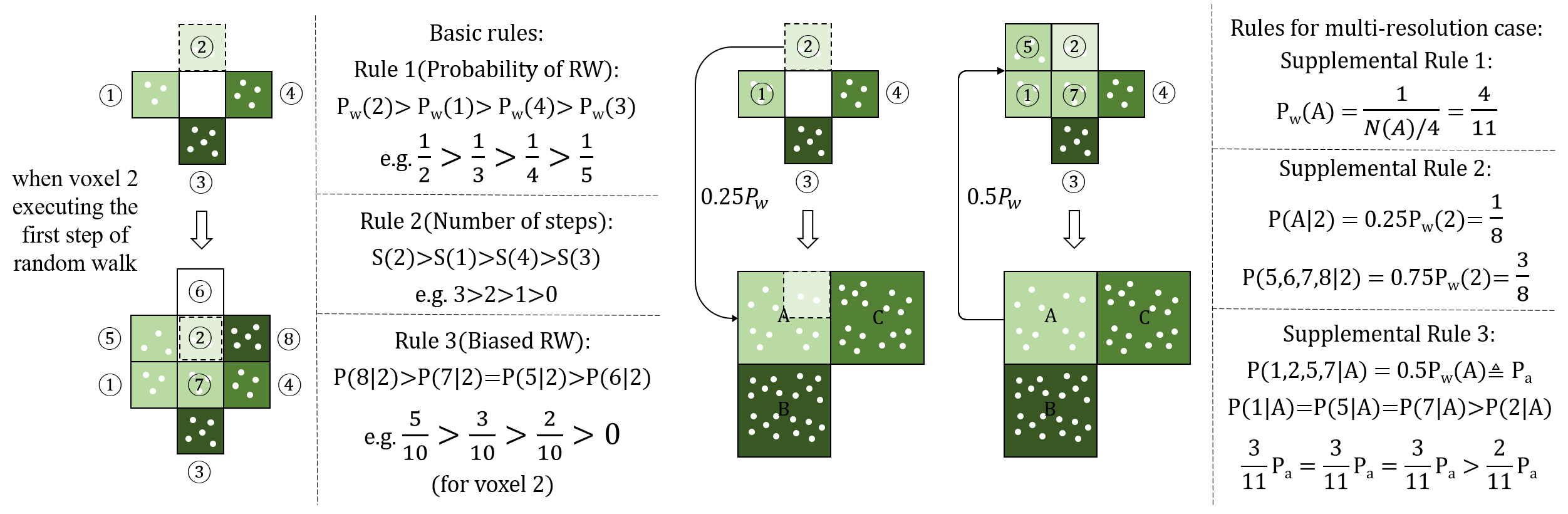}
    \vspace{-2.5ex}
    \caption{An example of 3 basic rules and 3 supplemental rules for biased random walk is shown above. Note that the basic rule 2 is only needed before starting random walk while all the other basic and supplemental rules are followed when executing every step of random walk}
    \label{fig: biasedRW}
    \vspace{-2.5ex}
\end{figure*}
As mentioned previously, we hope that by designing a biased random walk scheme, neighbor voxels will tend to move to areas with dense points. An intuitive idea is when a voxel contains fewer points, it should be more likely to execute random walk and take more steps on the same connected component. In addition, voxels should have a greater probability of transitioning to those with denser points. We formulate this idea as follows.



Suppose the j-th voxel contains $N(j)$ points, the maximum number of points is $n$, the probability of executing random walk is $P_w(j)$, the number of steps it takes is $S(j)$, the voxel index of the i-th step is $w_j(i)$, the set of four neighbor voxels of $w_j(i)$ is $V({w_j(i)})$, and the transition probability from i-th step voxel to the next step voxel is $P(w_j(i+1)|w_j(i))$, our mechanism is given by the following 3 basic rules:
\begin{small}
\begin{equation}
    P_w(j) = \frac{1}{N(j)}
\label{equ:cond1}
\end{equation}
\vspace{-1.0ex}
\begin{equation}
    S(j) = n - N(j)
\label{equ:cond2}
\end{equation}
\vspace{-2.0ex}
\begin{equation}
    P(w_j(i+1)|w_j(i))=\frac{N(w_j(i+1))}{\sum_{v\in V(w_j(i))}N(v)}
\label{equ:cond3}
\end{equation}
\end{small}
\vspace{-1.0ex}

where $P(w_j(i+1)|w_j(i))$ is not zero if and only if $w_j(i+1)$ and $w_j(i)$ are non-empty neighbor voxels to each other. From the first 2 rules, the more points a voxel has, the lower its random walk probability is and the fewer steps it takes. It should be noted that the number of steps are decided at the beginning for every neighbor voxel, which is different from the transition probability. In particular, when the number of points reaches the maximum, the step number is 0, meaning that once the random walking neighbor reaches the voxel with the largest number of points, it will not leave. Voxels with only one point take the most $n-1$ steps among all cases, and according to the statistics of random walk in 2D case, the distance traveled from starting point is approximately on the order of $\sqrt{n-1}$ on average. Finally, the third rule says when walking between voxels, the probability of transferring to voxels with dense points is higher, and the sum of probabilities is 1.

\begin{figure*}
\begin{center}
\includegraphics[width=1.0\linewidth]{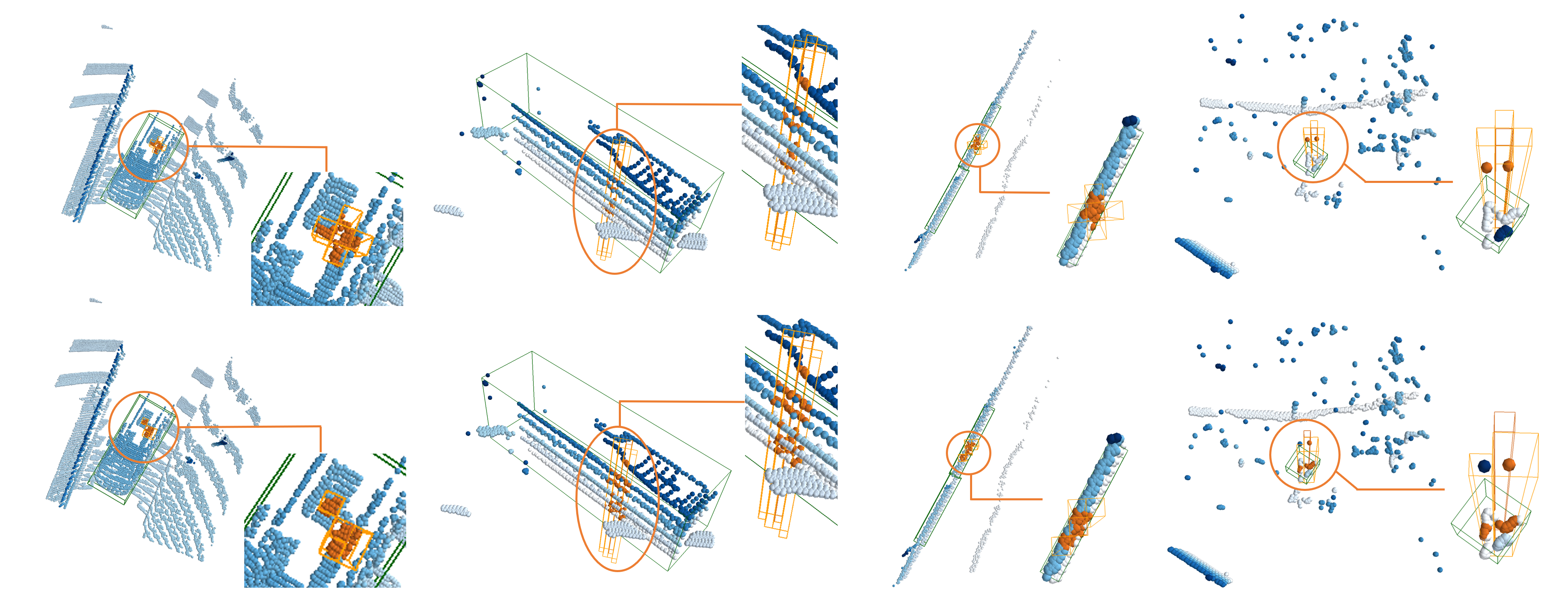}
\end{center}
\vspace{-2.5ex}
   \caption{Visualization of reconfigurable voxels. We show dilated (top) and reconfigurable voxels (bottom) in orange boxes. Points in the partitioned voxels are also marked in orange. For other points, the darker the blue, the higher the points on z axis. It demonstrates that reconfigurable voxels quantize the space adaptively, in which biased random walk helps neighbors cover more meaningful regions}
\label{fig: qualitative_RW}
\vspace{-2.5ex}
\end{figure*}

Up to now, we have preliminarily devised a scheme of biased random walk to achieve the transition between voxels that meet our requirements. It should be mentioned that this particular design sometimes needs to be adjusted according to the specific implementation and hyper parameters to ensure that voxel does not go too far. Specific adjustments are described in the appendix. See an example of this scheme in Fig.~\ref{fig: biasedRW} and voxel reconfiguration results in Fig.~\ref{fig: qualitative_RW}.

\subsection{Reconfigurable Voxels Encoder}\label{encoder}
With $n$ point features of the center voxel and $4n$ point features of neighbor voxels, we utilize a function, denoted as $\psi$, to extract voxel features. If the i-th input center voxel features and neighbor voxel features are denoted as $f_{v_i}$ and $f_{V(v_i)}$, the derived i-th voxel features is denoted as $F(v_i)$, then:
\vspace{-1ex}
\begin{small}
\begin{equation}
    F(v_i)=\psi(f_{v_i}, f_{V(v_i)})
\label{equ:encoder}
\end{equation}
\end{small}
where we take the original center as the center of reconfigurable voxels to obtain relative locations in $f_{V(v_i)}$. For SECOND and PointPillars, considering their different partition methods in $z$ axis, $\psi$ has different implementations. SECOND partitions the space more carefully, so it is more difficult to form a connected component. For instance, suppose a voxel at a certain height does not have non-empty neighbors in SECOND, the case can be not true if the neighbor pillar contains points at different heights in PointPillars. So it should be careful when we leverage neighbor voxel features in PointPillars: our encoder needs to ensure that neighbor pillar features will not overwhelm the original pillar information. To this end, we adopt $\psi$ as follows:
\begin{small}
\begin{equation}
    \psi(f_{v_i}, f_{V(v_i)}) = \phi_1([f_{v_i}, f_{V(v_i)}]_p)
\label{equ:SECOND_encoder}
\end{equation}
\vspace{-3ex}
\begin{equation}
    \psi(f_{v_i}, f_{V(v_i)}) = [\phi_2(f_{v_i}), \phi_2(\sum_{j=1}^{4} W_j(f_{v_i})f_{V_j(v_i)})]_f
\label{equ:Pillars_encoder}
\end{equation}
\end{small}
where $\phi_1$ is a low-level operation, average pooling, for SECOND, while $\phi_2$ is a high-level operation, shared MLP and maxpooling, for PointPillars. $W_j(f_{v_i})$ is the weight corresponding to the j-th neighbor of $v_i$, which is derived from $f_{v_i}$. In a word, we just encode the concatenated features (of different points) in SECOND, while concatenate the encoded features in PointPillars. From this perspective, our approach basically aggregates more meaningful point features locally for a better input representation, and thus ease the burden of learning a better $\psi$ as well as the following networks.

\subsection{Multi-resolution Reconfigurable Voxels}\label{multires}
So far, we have designed a method to construct reconfigurable voxels in the single-resolution case, in which we devise a scheme so-called \emph{intra-resolution} random walk. In order to make it more flexible and robust, we extend it to the case of multi-resolution random walk, namely, \emph{inter-resolution} random walk. Here we give a detailed implementation of two-resolution scenarios.

Firstly, suppose that under the initial resolution partition, the voxel size on the X-Y plane is $[l, w]$, and each voxel contains at most $n$ points. To preserve the resolution of the original voxel, we consider the second resolution with a larger-voxel partition: the voxel size on the X-Y plane is $[2l, 2w]$. Then a large voxel will contain up to 4 small voxels. In order to ensure the consistency of data format, we record the indices of 4 children voxels for the large ones when implementing voxel partition. After completing the partition, we randomly sample the points in the large voxel to make it contain up to $n$ points. As a result, the voxels with dense points will not contain more points with the change of spatial  quantization, whereas the voxels with less than $n$ points have chance containing enough data. Besides, this design also facilitates the convenience of subsequent voxel feature extraction.

Problems mentioned in Sec.~\ref{sec:biasedRW} also exist when it comes to random walk operations between different resolutions. As Fig.~\ref{fig: biasedRW} shows, we put forward 3 supplemental rules for multi-resolution case. Firstly, when computing $P_w$, we need to divide the number of points by 4 to make it consisent with the single-resolution case. For supplemental rule 2 and 3, we assume that the transition probability from smaller voxel to larger one is $0.25P_w$, and from larger voxel to smaller one is $0.5P_w$, which ensures that all voxels will remain in the original resolution at a higher probability. Note that it follows similar rules as the basic rule 3 when choosing which small voxel to transition. Finally, we will also record the neighbors of large voxels, and it satisfies Eqn.~\ref{equ:cond3} when they execute random walk in the graph composed of large voxels.

Thus, we complete the generalization to the multi-resolution case. Specification of the algorithm is included in the appendix. In conclusion, this extension makes reconfiguration more flexible. In particular when the points in a voxel are very sparse, the higher probability to be a larger voxel will make it easier to contain more points, so as to ease the difficulty caused by sparsity. It should be noted that our purpose is to construct the new representation with voxels in different resolutions given the local spatial distribution of point clouds, which is different from general multi-scale tricks.
\section{Experimental Setup}
\subsection{Datasets \& Evaluation Metrics} We evaluated our approach on three commonly used benchmarks: nuScenes~\cite{nuScenes}, Lyft~\cite{Lyft} and KITTI~\cite{KITTI}. NuScenes dataset is split in 700/150/150 scenes for training/validation/testing respectively. There are overall 1.4M annotated 3D boxes, far more than KITTI's 200K 3D boxes in 22 scenes. Lyft dataset has 180 and 218 scenes for training and testing respectively. It can be seen that nuScenes and Lyft have more data, more object categories and richer scenes than KITTI. Therefore, at first, we conducted toy experiments on KITTI to analyze the computational complexity and the efficacy of our method under different settings. Then we designed experiments on nuScenes and Lyft to test it on large-scale datasets. Finally, more detailed ablation studies on KITTI are given. It should be noted that nuScenes and Lyft have the same data format, and need to predict one key frame detection result every ten frames. Therefore, in those experiments, we transformed the point clouds of ten consecutive frames into the coordinate system of key frames and input them to the network for detection. As for metrics, distance-based mAP and nuScenes detection score (NDS\footnote{NDS is a more comprehensive metric with consideration of attribute and velocity prediction in \cite{nuScenes}.}) were used as the main metrics on nuScenes, while mAP of all categories was compared under 0.5-0.95 IOU on Lyft. Here we name the much more strict metric in Lyft as mAP-3D for clarification. We follow the official evaluation protocol in KITTI experiments as well, \emph{i.e.}, mAP was compared for different categories with 0.7 IOU threshold for car and 0.5 IOU for pedestrian and cyclist. 

\subsection{Implementation Details} \label{sec:details}
\noindent\textbf{Network Architectures}~~ Our whole framework follows the ideas of PointPillars and SECOND with the following adjustments in specific details.

First, when extracting features from voxels, we use different point features and different settings of X-Y resolution, max number of voxels and max number of points per voxel for different experiments.
Another change on the PointPillars is that we implement multi-group head for the experiments on nuScenes and Lyft given the category diversity.
See more details in the appendix.



\noindent\textbf{Loss}~~ We use a loss function similar to that described in \cite{PointPillars,SECOND}. It should be noted that we need to predict the object's velocity and attribute in the nuScenes experiment, so we add the velocity into the regression target and add attribute classification loss into the overall loss.
\begin{small}
\begin{equation}
    L_{loc}=\sum_{b\in(x,y,z,w,l,h,\theta,v_x,v_y)}\text{SmoothL1}(\Delta b)
\end{equation}
\end{small}
where the weight of $x$, $y$, $z$, $w$, $l$, $h$, $\theta$ error is 1 and the weight for $v_x$, $v_y$ is 0.5. The total loss is:
\begin{small}
\begin{equation}
    L=\frac{1}{N_{pos}}(\beta_{loc}L_{loc}+\beta_{cls}L_{cls}+\beta_{attr}L_{attr}+\beta_{dir}L_{dir})
\end{equation}
\end{small}
where $N_{pos}$ is the number of positive anchors and $\beta_{loc}=2$, $\beta_{cls}=1$, $\beta_{attr}=1$ and $\beta_{dir}=0.2$.

\noindent\textbf{Training Parameters}~~ For all the experiments, we trained randomly initialized networks end-to-end. Models were trained with ADAM optimizer~\cite{ADAM}, in which we adopted one-cycle policy~\cite{1cycle}.

\noindent\textbf{Data Augmentation}~~ Data augmentation is particularly important for 3D detection.
First, we establish the ground truth database of all objects as mentioned in ~\cite{SECOND}. During training, we sample a few objects which have fewer instances, and place them into different point clouds. Because this kind of augmentation may be unreasonable due to the characteristic of LiDAR sampling, we also analyze the number of different categories of objects in all samples, select specific samples, copy them, and alleviate the imbalance of the number of objects in all categories as \cite{ClsBalanced} proposed. Finally, we randomly flip the LiDAR sweep along the x-axis or y-axis to realize global augmentation.

\begin{table}
\scriptsize
\begin{minipage}[c]{.3\linewidth}
    \begin{center}
    \caption{Inference speed of models with and without reconfigurable voxels.}
    \vspace{-1ex}
    \label{tab:realtime}
    \begin{tabular}{c|c}
    \hline
    \textbf{Method} & \textbf{Speed(Hz)} \\
    \hline\hline
    SECOND & 23 \\
    \hline
    Reconfig SECOND & 21 \\
    \hline
    PointPillars & 53 \\
    \hline
    Reconfig PP & 47 \\
    \hline
    \end{tabular}
    \end{center}
\end{minipage}
\hspace{2mm}
\begin{minipage}[c]{.66\linewidth}
    \begin{center}
    \caption{Results in different distance ranges on the nuScenes val benchmark, where the object distance from ego vehicle is denoted as $d$ and \emph{nuScenes range} refers to the official evaluation range}
    \vspace{-1ex}
    \label{tab:nuScenes_dist}
    \begin{tabular}{c|c|c|c|c|c|c}
    \hline
    \multirow{2}*{Method} & \multicolumn{2}{c|}{d $<$ 20m} & \multicolumn{2}{c|}{d $\ge$ 20m} & \multicolumn{2}{c}{nuScenes range}\\
    \cline{2-7}
    ~ & mAP & NDS & mAP & NDS & mAP & NDS\\
    \hline\hline
    PointPillars & 45.3 & 58.1 & 11.8 & 33.8 & 30.3 & 48.6\\
    \hline
    Reconfig PP (sing-res) & \textbf{48.8} & \textbf{60.3} & 12.4 & \textbf{38.2} & 32.8 & 50.3\\
    \hline
    Reconfig PP (multi-res) & 48.4 & 59.7 & \textbf{12.6} & \textbf{38.2} & \textbf{32.9} & \textbf{50.5}\\
    \hline
    \end{tabular}
    \end{center}
\end{minipage}
\vspace{-1.5ex}
\end{table}

\begin{table}
\scriptsize
\caption{Distance-based mAP by categories compared to PointPillars on the nuScenes test 3D detection benchmark. Here according to the average size of all the bounding boxes, we consider the first 5 categories (car, bus, truck, trailer and construction vehicle) as large objects while the last 5 categories (pedestrian, barrier, traffic cone, motorcycle and bicycle) as small objects. We compute the mAP of all the small objects and record it as mSAP in the table}
\label{tab:nuScenes}
\vspace{-0.3cm}
\begin{center}
\begin{tabular}{c|c|c|c|c|c|c|c|c|c|c|c|c}
\hline
Method & Car & Bus & Truck & Trail & CV & Ped & Bar & TC & Moto & Bicy & \textbf{mAP} & \textbf{mSAP}\\
\hline\hline
PointPillars & 74.4 & 38.5 & 23.4 & 36.1 & 4.8 & 60.1 & 30.5 & 19.8 & 12.9 & 0.1 & 30.1 & 24.7\\
\hline
Reconfig PP (sing-res) & 75.6 & 38.5 & 26.5 & \textbf{38.9} & \textbf{7.5} & \textbf{63.1} & 34.4 & 23.8 & \textbf{15.2} & 0.1 & 32.4 & 27.3\\
\hline
Reconfig PP (multi-res) & \textbf{75.8} & \textbf{39.5} & \textbf{27.2} & 38.0 & 6.5 & 62.5 & \textbf{34.9} & \textbf{25.7} & \textbf{15.2} & \textbf{0.2} & \textbf{32.5} & \textbf{27.7}\\
\hline
\end{tabular}
\end{center}
\vspace{-5.5ex}
\end{table}

\section{Results}
In this section, we first present the complexity analysis of our reconfiguration algorithm along with relevant experimental results. Then quantitative and qualitative results are given to show the performance improvement, especially the performance for small and distant objects. For fairness, all of the following experiments are conducted without further tuning network architecture and parameters or introducing more tricks.
\subsection{Complexity Analysis}
Firstly, let us briefly compare the complexity of vanilla voxelization and our improved version. Suppose there are $N$ points and $M$ voxels, the reconfiguration process only adds constant operations when traversing all points, as well as one-time traversal of voxels when performing random walk. So the complexity changes from $O(N)$ to $O(N+M)$. The more points each voxel contains, the greater the ratio $\frac{N}{M}$ is, then the effect on the efficiency of voxelization is more limited. 

To indicate the influence of this representation on the inference speed more empirically, we validate it in KITTI experiments (Tab.~\ref{tab:realtime}). Our method hardly affects the algorithm efficiency and the inference speed is still much faster than point-based methods (about 10Hz of \cite{PointRCNN,STD,FastPointRCNN}) and can achieve real-time detection.

\subsection{Quantitative Analysis}
\textbf{Toy experiments on KITTI}\quad First, we did a series of preliminary experiments on the KITTI dataset to investigate the effectiveness of our method under different settings. As shown in Fig.~\ref{fig: teaser}, taking the representative small object, pedestrian, as an example, we find that our method can consistently improve the detection performance when using different pillar or voxel resolutions. In addition, we also compare their performance at different distances in the experiments where minimum pillar or voxel resolution is adopted. As we expected, performance improvements become more evident as distance increases. See more detailed results in the appendix.

\noindent\textbf{Experiments on large-scale datasets}\quad Then we test our methods on large-scale datasets. Considering the large amount of data and the difficulty of training networks including SECOND, we only give the experimental results on PointPillars here. Due to higher ranked models on these two benchmarks typically adopt heavy heads, we validate the efficacy of our methods both on lightweight, real-time baselines (Fast PP) and those with higher performance (Heavy PP).

Firstly, in order to study the improvement details, we evaluate the detection performance of objects from different categories and distance ranges, where the latter is conducted on the validation set. Taking the Fast PP experiments as the example, from Tab.~\ref{tab:nuScenes}, it can be seen that mAPs of smaller objects are greatly improved, among which the multi-resolution version increases mAPs of pedestrian, barrier, traffic cone and motorcycle by 2.4\%, 4.4\%, 5.9\% and 2.3\% respectively. In addition, from Tab.~\ref{tab:nuScenes_dist}, we can observe that in terms of distant object detection in the distance range over 20m, NDS is increased by up to 4.4\%. Meanwhile, in the above two experiments, the detection performance of large and close objects is not affected, but most aspects are also improved. Finally, compared with baseline models, our method can respectively improve 2.4\% mAP, 2.1\% NDS and 2.3\% mAP, 2.2\% NDS on top of Fast PP and Heavy PP. With further training steps and adding more data augmentation (without model ensemble), we achieve 48.5\% mAP and 59.0\% NDS in our final model, which is comparable with ensembled top entries~\cite{ClsBalanced} and outperforms all the published methods.

\begin{table}
\tiny
\begin{minipage}[t]{.45\linewidth}
    \centering
    \caption{Results on the nuScenes dataset}
    \vspace{-1ex}
    \label{tab:nuScenes_SOTA}
    \begin{tabular}{c|c|c|c}
        \hline
        Method & Modality & mAP & NDS\\
        \hline
        MAIR~\cite{MAIR} & RGB & 30.4 & 38.4\\
        Freespace~\cite{WYSIWYG} & LiDAR & 35.0 & 41.9\\
        PP~\cite{PointPillars} & LiDAR & 30.5 & 45.3\\
        SECOND~\cite{SECOND} & LiDAR & 31.6 & 46.8\\
        SHAPNET~\cite{SHAPNET} & LiDAR & 32.4 & 48.4\\
        3DSSD~\cite{3DSSD} & LiDAR & 42.6 & 56.4\\
        Painting~\cite{Painting} & LiDAR+RGB & 46.4 & 58.1\\
        CBGS~\cite{ClsBalanced} & LiDAR & \textbf{52.8} & \textbf{63.3}\\
        \hline
        Fast PP~\cite{PointPillars} & LiDAR & 30.1 & 48.5\\
        +Reconfig & LiDAR & 32.4 & 50.2\\
        +Multi-res & LiDAR & \textbf{32.5} & \textbf{50.6}\\
        \hline
        Heavy PP & LiDAR & 43.4 & 54.1\\
        +Reconfig & LiDAR & 45.4 & 56.1\\
        +Multi-res & LiDAR & \textbf{45.7} & \textbf{56.3}\\
        \hline
        Ours (Final) & LiDAR & 48.5 & 59.0 \\
        \hline
    \end{tabular}
    \vspace{-5.5ex}
\end{minipage}
\hspace{1mm}
\begin{minipage}[t]{.55\linewidth}
    \begin{center}
    \caption{Results on the Lyft dataset}
    \vspace{-2ex}
    \label{tab:Lyft_SOTA}
    \begin{tabular}{c|c|c|c}
        \hline
        Team/Method & Reference & Modality & mAP-3D\\
        \hline
        STL-IV Lab & 11st place & -  & 14.2\\
        MIT HAN Lab & 10th place & -  & 14.4\\
        ... & ... & - & -\\
        Wenjing (single model) & 1st place & LiDAR & \textbf{17.9}\\
        \hline
        VoxelNet~\cite{VoxelNet} & CVPR 2018 & LiDAR & 10.1\\
        SECOND~\cite{SECOND} & Sensors 2018 & LiDAR & 13.0\\
        \hline
        Fast PP~\cite{PointPillars} & CVPR 2019 & LiDAR & 10.4\\
        +Reconfig & - & LiDAR & 11.3\\
        +Multi-res & - & LiDAR & \textbf{11.4}\\
        \hline
        Heavy PP & CVPR 2019 & LiDAR & 11.9\\
        +Reconfig & - & LiDAR & 12.7\\
        +Multi-res & - & LiDAR & \textbf{12.9}\\
        \hline
        Larger range & CVPR 2019 & LiDAR & 16.0\\
        +Reconfig & - & LiDAR & 16.7\\
        +Multi-res & - & LiDAR & \textbf{16.9}\\
        \hline
    \end{tabular}
    \end{center}
    \vspace{-5.5ex}
\end{minipage}
\end{table}

In addition to nuScenes, we also tested on Lyft benchmark as Tab.~\ref{tab:Lyft_SOTA} shows, where the \emph{Larger range} refers to the change of x,y range both from [-49.6, 49.6] to [-89.6, 89.6] on the basis of Heavy PP. Our final model can consistently achieve about 1.0\% mAP increase for all 3 baselines under the more difficult mAP-3D metric. Furthermore, this improvement is mainly achieved by the enhanced detection performance of small and distant objects, which are only a minority of all the objects. Detailed analysis of Lyft results can be referred to the appendix.

\noindent\textbf{Ablation studies}\quad Finally, we take SECOND experiments on KITTI as the example to give more detailed ablation studies. In the experiments, we controlled whether to add 4 neighbor voxels (Dilated, abbrev. DL in Tab.~\ref{tab:KITTI}), whether to reconfigure sparse voxels, whether to reconfigure dense voxels, whether in different resolutions, and carried out the corresponding experiments. Here \emph{dense voxels} means that they contain the maximum number of points while \emph{sparse} indicates otherwise, and \emph{DL} corresponds to the case with the same framework but  without neighbor voxels reconfiguration (see the comparison in Fig.~\ref{fig: teaser}). It turned out that the improvement of detecting larger objects like car is slight but stable. On cyclist and pedestrian, almost all of our models are better than the baseline model, which shows the necessity of improving the representation. Especially for cyclist, our best model can achieve better mAPs on the easy, moderate and hard sets by 3.43\%, 5.87\% and 1.28\% increase respectively. Most importantly, comparison with the dilated voxels (DL) based on the original voxel neighbor layout shows the effectiveness of our reconfiguration mechanism.
\begin{table*}
\scriptsize
\begin{center}
\caption{Ablation studies on the KITTI val 3D detection benchmark}
\vspace{-1ex}
\label{tab:KITTI}
\begin{tabular}{c|c|c|c|c|c|c|c|c|c|c|c|c|c}
\hline
\multirow{2}*{DL} & \multirow{2}*{\shortstack{Sparse\\Reconfig}} & \multirow{2}*{\shortstack{Dense\\Reconfig}} & \multirow{2}*{\shortstack{Multi\\res}} & \multirow{2}*{\textbf{mAP}} & \multicolumn{3}{c|}{Car} & \multicolumn{3}{c|}{Cyclist} & \multicolumn{3}{c}{Pedestrian}\\
\cline{6-14}
~ & ~ & ~ & ~ & ~ & Easy & Mod. & Hard & Easy & Mod. & Hard & Easy & Mod. & Hard\\
\hline\hline
$\times$ & $\times$ & $\times$ & $\times$ & 66.76 & 88.31 & 77.79 & 75.91 & 77.07 & 59.95 & 58.96 & 59.78 & 53.22 & 49.88\\
\hline
$\surd$ & $\times$ & $\times$ & $\times$ & 67.07 & 88.39 & 77.90 & 75.92 & 75.17 & 58.94 & 57.49 & 61.39 & 57.15 & 51.31\\
\hline
$\surd$ & $\surd$ & $\times$ & $\times$ & 67.08 & 88.17 & 77.38 & 75.56 & 77.48 & 59.94 & 58.02 & 60.64 & 56.33 & 50.22\\
\hline
$\surd$ & $\times$ & $\surd$ & $\times$ & 67.40 & 88.14 & 77.75 & 76.03 & 76.36 & 60.03 & 57.71 & 61.70 & 57.71 & 51.13\\
\hline
$\surd$ & $\surd$ & $\surd$ & $\times$ & 68.36 & \textbf{88.88} & 78.09 & 76.13 & 79.63 & 61.87 & 59.26 & \textbf{61.97} & \textbf{57.77} & \textbf{51.63}\\
\hline
$\surd$ & $\surd$ & $\surd$ & $\surd$ & \textbf{68.41} & 88.65 & \textbf{78.22} & \textbf{76.21} & \textbf{80.50} & \textbf{65.82} & \textbf{60.24} & 61.63 & 54.08 & 50.33\\
\hline
\end{tabular}
\end{center}
\vspace{-6.5ex}
\end{table*}
\subsection{Qualitative Analysis}
We visualize some samples to show the results of voxel reconfiguration (Fig.~\ref{fig: qualitative_RW}). It can be seen that with the help of our mechanism, neighbor voxels move to regions with more points and implicitly follow surface and shape of objects as well. We thus believe that voxel encoder can benefit a lot from this more reasonable spatial quantization. See the appendix for qualitative analysis of detection results on nuScenes.

\vspace{-1ex}
\section{Conclusion}
\label{sec:conclusion}
In this paper, we propose \emph{Reconfigurable Voxels}, a novel representation that can significantly improve the imbalance of sampling points in different voxels caused by sparsity and irregularity of LiDAR point cloud. We demonstrate that on various 3D detection benchmarks, incorporating this lightweight representation into the state-of-the-art voxel-based frameworks can greatly enhance the performance in terms of small and distant objects without much computation overhead. Future work includes designing this mechanism more carefully and figuring out this problem in point-based and multi-sensor fusion methods. 



\acknowledgments{This work is partially supported by the SenseTime Collaborative Grant on Large-scale Multi-modality Analysis (CUHK Agreement No. TS1610626 \&  No. TS1712093), the General Research Fund (GRF) of Hong Kong (No. 14236516 \&  No. 14203518).}


\bibliography{ref}  

\clearpage
\begin{center}
    \Large
    \textbf{Appendix}
\end{center}

\setcounter{section}{0}
\section{Algorithm Specification}
Here we give the specification for the algorithm of contructing reconfigurable voxels in the multi-resolution case (Algorithm~\ref{alg:ReconfigPartition}). To sum up, we can complete this in one traversal of all point clouds for every sample: for every point, locate its coordinates in two-resolution voxel maps at first and record the adjacency of relevant voxels; then for all neighbor voxels, compute transition distributions and implement reconfiguration; finally, resample points in larger voxels.

\begin{algorithm}
  \caption{Multi-resolution Reconfig. Voxel Partition}
  \label{alg:ReconfigPartition}
  \begin{algorithmic}[1]
    \Require
        (1) point cloud data $P=\{p_i, i=1,...,n\}$;
        (2) maximum number of points per voxel $m$;
        (3) maximum number of voxels $N$ (in resolution 1);
    \Ensure
        (1) voxel coordinates of two resolutions ($C_1$, $C_2$); 
        (2) voxel features of two resolutions ($F_1$, $F_2$); 
        (3) number of points in voxels of two resolutions ($N_1$, $N_2$); 
        (4) graph $G$ (to record indices and resolutions of voxel neighbors);
    \For{every point $p_i$}
    	\If{$p_i$ is not in the detection range}
    		\State continue;
    	\EndIf
    	\State $//$ Denote variables of smaller voxel with index 1 
    	\State $//$ and larger one with index 2
    	\State Locate its voxel coordinates $c_1$ in resolution 1;
    	\State Locate its voxel coordinates $c_2$ in resolution 2;
    	\If{voxel1 at $c_1$ not recorded yet}
    		\If{number of recorded voxels1 $\ge N$}
    			\State break;
    		\EndIf
    		\State create a new voxel1;
    		\If{voxel2 at $c_2$ not recorded yet}
    			\State create a new voxel2;
    			\If{left/right/back/front neighbor exists}
    				\State record their adjacency (in resolution2);
    			\EndIf
    		\EndIf
    		\State record the parent index for voxel1;
    		\State record the children index for voxel2;
    		\If{left/right/back/front neighbor exists}
    			\State record their adjacency (in resolution1);
    		\EndIf
    	\EndIf
    	\If{number of points in that voxel $<m$}
    		\State add the point features of voxel1 into $F_1$;
    	\EndIf
    \EndFor
    \State compute transition distribution probabilities according to number of points in voxels;
    \For{every voxel1 $v_i$}
    	\For{every neighbor of $v_i$}
    		\State compute number of steps $S$;
    		\State conduct random walk for $S$ times;
    	\EndFor
    	\State record final adjacency into $G$;
    \EndFor
    \State resample points of voxels2 (up to $m$ points per voxel) and record them into $F_2$
  \end{algorithmic}
\end{algorithm}

\section{Supplementary Results}
\subsection{Quantitative Results}
In this section, we present more quantitative analysis on Lyft and detailed results compared with other state-of-the-art methods on KITTI.

First, given the similarity of data format on nuScenes~\cite{nuScenes} and Lyft~\cite{Lyft}, it is convenient to compare performance of different methods on the Lyft under nuScenes metrics. Considering that we do not need to predict velocity and attribute on the Lyft, NDS is not suitable for evaluation. So we test their distance-based mAPs by categories on our split Lyft validation set, where \emph{v1} and \emph{v2} represent different feature fusion methods in Reconfigurable PointPillars.\footnote{To use our metrics, we split the official training data into train/val set by the same ratio as nuScenes.} See more details in the Sec. 4 of appendix.

Results of 7 categories based on our reproduced Fast PointPillars are shown in Tab.~\ref{tab:Lyft}. Two other categories, animal and emergency vehicle, are not shown here because their training data is too limited (only 186 animal instances and 132 emergency instances compared with 534911 cars). It can be seen that our representation can greatly enhance the detection performance, especially for small objects. For example, our model can increase mAPs of pedestrian and bicycle by up to 5.7\% and 13.6\%. Note that motorcycle also has limited 818 instances, so the improvement is not much noteworthy. Finally, the original model can be improved by 3.2\% mAP for all objects and 5.7\% mAP for small objects. It further demonstrates the efficacy of our representation.

Then detailed results on the KITTI benchmark are shown in Tab.~\ref{tab:KITTI_SOTA}, which lists several methods using point clouds as input, including multi-modal data fusion methods, point-based methods and voxel-based methods. Note that only most of the published methods which provide test results for all 3 types of objects are listed here. In the same way, our method improves the detection performance of cyclist and pedestrian. The improvement for PointPillars is mainly focused on the pedestrian. We speculate that there may be differences in the network details between the baseline and the original paper. In fact, the reproduced baseline cannot fully achieve the decent performance of cyclist as the original paper. In addition, we find that for the detection of multi-class objects, there are certain mutual constraints between different categories, so maybe the performance improvement of cyclist and pedestrian can maintain a certain level together, but one can influence the other. The specific solution may include adopting different detection heads, which will be our future work.

We can also see that point-based methods dominate the KITTI benchmark in terms of performance currently. So the trade-off between performance and efficiency as well as how to address the irregularity distribution problem of LiDAR data in point-based methods are worthy of further study. 

\begin{table*}
\scriptsize
\begin{center}
\caption{Distance-based mAP by categories compared to PointPillars on the Lyft val 3D detection benchmark. Note that the results here are from the experiment with Fast PP as baseline in the paper. According to the average size of all the bounding boxes, we consider the first 4 categories (car, other vehicle, bus and truck) as large objects while the last 3 categories (pedestrian, bicycle and motorcycle) as small objects. We compute the mAP of all the small objects and record it as mSAP in the table}
\vspace{2ex}
\label{tab:Lyft}
\begin{tabular}{c|c|c|c|c|c|c|c|c|c}
\hline
Method & Car & Other Veh. & Bus & Truck & Ped & Bicycle & Moto & \textbf{mAP} & \textbf{mSAP}\\
\hline\hline
PointPillars & 93.5 & 63.1 & \textbf{46.9} & 43.1 & 48.1 & 37.4 & 3.9 & 48.0 & 29.8\\
\hline
Reconfig PointPillars (sing-res v1) & 93.7 & 65.6 & 44.1 & 46.1 & \textbf{53.8} & 46.6 & 2.1 & 50.3 & 34.2\\
\hline
Reconfig PointPillars (sing-res v2) & 92.8 & 61.5 & 42.1 & 42.7 & 50.9 & 47.4 & \textbf{5.6} & 49.0 & 34.6\\
\hline
Reconfig PointPillars (multi-res) & \textbf{93.8} & \textbf{65.7} & 45.7 & \textbf{46.8} & 52.4 & \textbf{51.0} & 3.1 & \textbf{51.2} & \textbf{35.5}\\
\hline
\end{tabular}
\end{center}
\vspace{-4ex}
\end{table*}

\begin{table*}
\tiny
\begin{center}
\caption{Results on the KITTI dataset}
\label{tab:KITTI_SOTA}
\begin{tabular}{c|c|c|c|c|c|c|c|c|c|c|c|c}
\hline
 \multirow{2}*{Method} & \multirow{2}*{Reference} &
 \multirow{2}*{Modality} & \multirow{2}*{\textbf{mAP}} & \multicolumn{3}{c|}{Car} & \multicolumn{3}{c|}{Cyclist} & \multicolumn{3}{c}{Pedestrian}\\
\cline{5-13}
~ & ~ & ~ & ~ & Easy & Mod. & Hard & Easy & Mod. & Hard & Easy & Mod. & Hard\\
\hline
AVOD-FPN~\cite{AVOD} & IROS 2018 & LiDAR+RGB & 56.84 & 83.07 & 71.76 & 65.73 & 63.76 & 50.55 & 44.93 & 50.46 & 42.27 &	39.04\\
F-PointNet~\cite{F-PointNet} & CVPR 2018 & LiDAR+RGB & 57.86 & 82.19 & 69.79 & 60.59 & 72.27 & 56.12 & 49.01 & 50.53 & 42.15 & 38.08\\
F-ConvNet~\cite{F-ConvNet} & IROS 2019 & LiDAR+RGB & 63.15 & 87.36 & 76.39 & 66.69 & \textbf{81.98} & \textbf{65.07} & 56.54 & 52.16 & \textbf{43.38} & 38.80\\ 		
\hline
PointRCNN~\cite{PointRCNN} & CVPR 2019 & LiDAR & 60.33 & 86.96 & 75.64 & 70.70 & 74.96 & 58.82 & 52.53 & 47.98 & 39.37 & 36.01\\
STD~\cite{STD} & ICCV 2019 & LiDAR & 63.60 & \textbf{87.95} & \textbf{79.71} & \textbf{75.09} & 78.69 & 61.59 & 55.30 & \textbf{53.29} & 42.47 & 38.35\\
\hline\hline
VoxelNet~\cite{VoxelNet} & CVPR 2018 & LiDAR & 50.99 & 77.47 & 65.11 & 57.73 & 61.22 & 48.36 & 44.37 & 39.48 & 33.69 & 31.51\\
Part A2~\cite{PartA2} & TPAMI 2020 & LiDAR & \textbf{63.99} & 87.81 & 78.49 & 73.51 & 79.17 & 63.52 & \textbf{56.93} & 53.10 & 43.35 & \textbf{40.06}\\
\hline
SECOND~\cite{SECOND} & Sensors 2018 & LiDAR & 58.35 & 83.13 & 73.66 & 66.20 & 70.51 & 53.85 & 46.90 & 51.07 & 42.56 & 37.29\\
+Reconfig & - & LiDAR & +1.31 & +0.88 & -0.33 & +1.53 & +1.08 & +2.00 & +2.68 & +1.05 & +1.14 & +1.75\\
\hline
PointPillars~\cite{PointPillars} & CVPR 2019 & LiDAR & 60.80 & 82.58 & 74.31 & 68.99 & 77.10 & 58.65 & 51.92 & 51.45 & 41.92 & 38.89\\
+Reconfig & - & LiDAR & +0.68 & +0.78 & +0.21 & +0.29 & +0.43 & -0.23 & +0.27 & +1.80 & +1.4 & +1.14\\
\hline
\end{tabular}
\end{center}
\vspace{-5.5ex}
\end{table*}

\subsection{Qualitative Results}
In this section, we give some examples of detection results on nuScenes. Through these examples, we can intuitively observe the detection results and see the improvement of our model in detecting small and distant objects.

\begin{figure*}
\begin{center}
\includegraphics[width=1.0\linewidth]{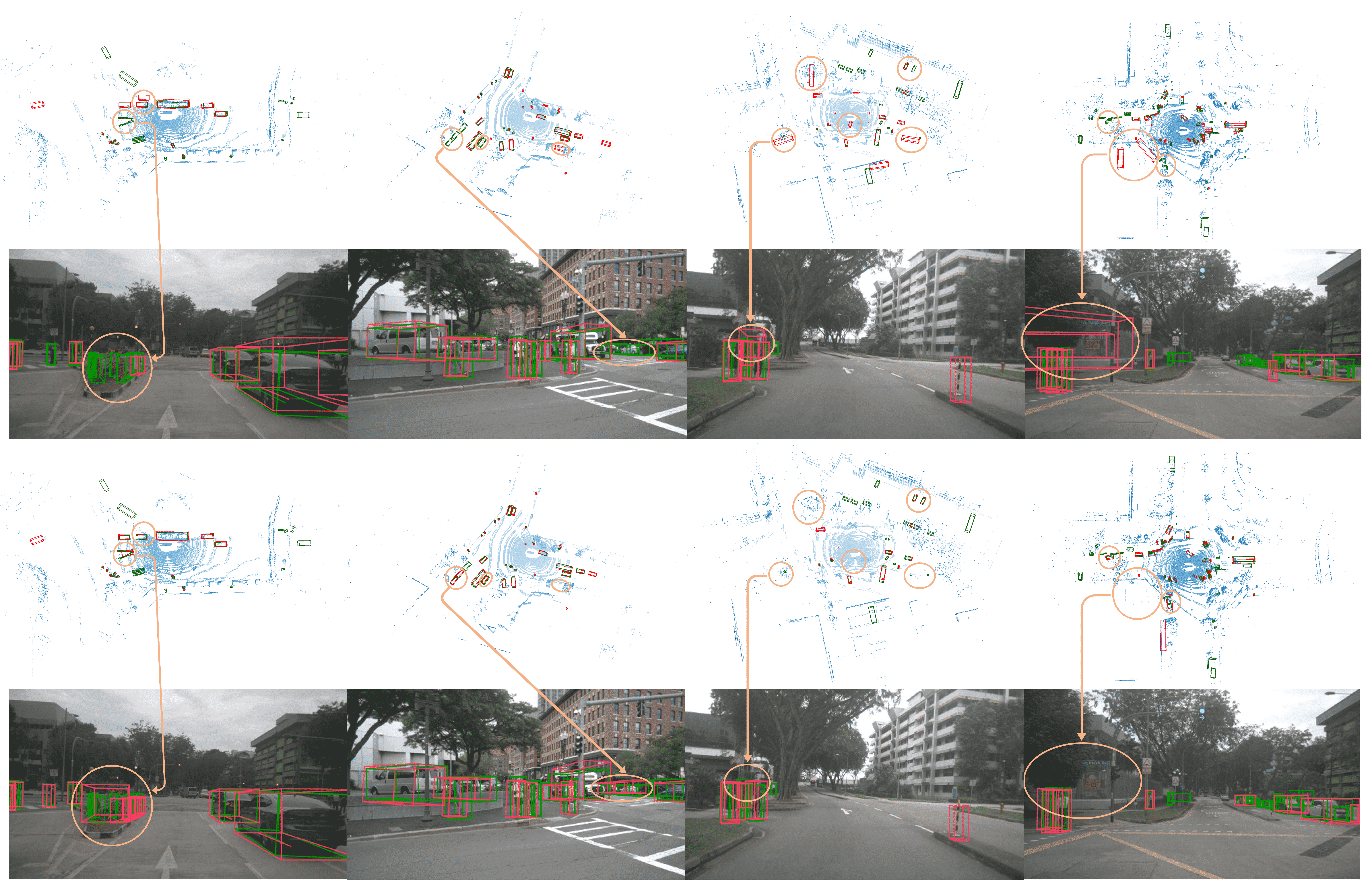}
\end{center}
  \caption{Qualitative analysis of nuScenes results. We show 3D bounding boxes, predicted results in red and ground truth in green, both in LiDAR point cloud and their projection into the image for visualization. The top 2 and bottom 2 rows are the results of our baseline and the model improved by reconfigurable pillars respectively. Less false and more correct detection of small and distant objects shows the improvement, which are marked with orange circles. Note that apart from the objects can be seen in images, there are more samples marked with orange circles in bird view}
\label{fig: qualitative}
\end{figure*}

In Fig.~\ref{fig: qualitative}, the near barriers in the first group of samples and the far-away little occluded cars in the second group of samples shows the improvement when detecting small and distant objects, while the last 2 groups of samples show that the improved model reduces false positive detections of large objects in the distance and small objects in the near. 

Besides, we can see some interesting phenomena from the failure examples in these results. For instance, the vehicle detected by mistake in the last sample is closely related to the roadside building, which indicates that the detector sometimes cannot distinguish the corner of car and building. In the third sample, both models detect an obstacle that was not annotated.

\section{Detailed Analysis of Change in Distribution}
As mentioned in the paper, our reconfigurable voxels effectively mitigate the difficulty caused by sparsity and irregularity, which can be reflected in the more balanced distribution with respect to number of points per voxel. So in this section, we give quantitative and qualitative results in this respect.

\subsection{Quantitative Analysis}
To make this result more convincing, we conduct the study on the validation set of nuScenes instead of just on several samples. Considering the number of points involved in every voxel can be regarded as a kind of \emph{ratio scale} in some sense, we use the \emph{coefficient of variation} as the indicator to measure their dispersion of the distribution. Tab.~\ref{tab:CV_analysis} shows the result, which is consistent with our observation on the qualitative results.

\begin{table}
\scriptsize
\begin{center}
\caption{Coefficient of variation w.r.t. points in original voxels and reconfigurable voxels}
\label{tab:CV_analysis}
\begin{tabular}{c|c}
\hline
\textbf{Method} & \textbf{Coefficient of Variation} \\
\hline\hline
PointPillars & 0.9766 \\
\hline
Reconfig PointPillars (sing-res) & 0.7695 \\
\hline
Reconfig PointPillars (multi-res) & \textbf{0.6796} \\
\hline
\end{tabular}
\end{center}
\vspace{-3.5ex}
\end{table}

\subsection{Qualitative Analysis}
\begin{figure*}
\begin{center}
\includegraphics[width=1.0\linewidth]{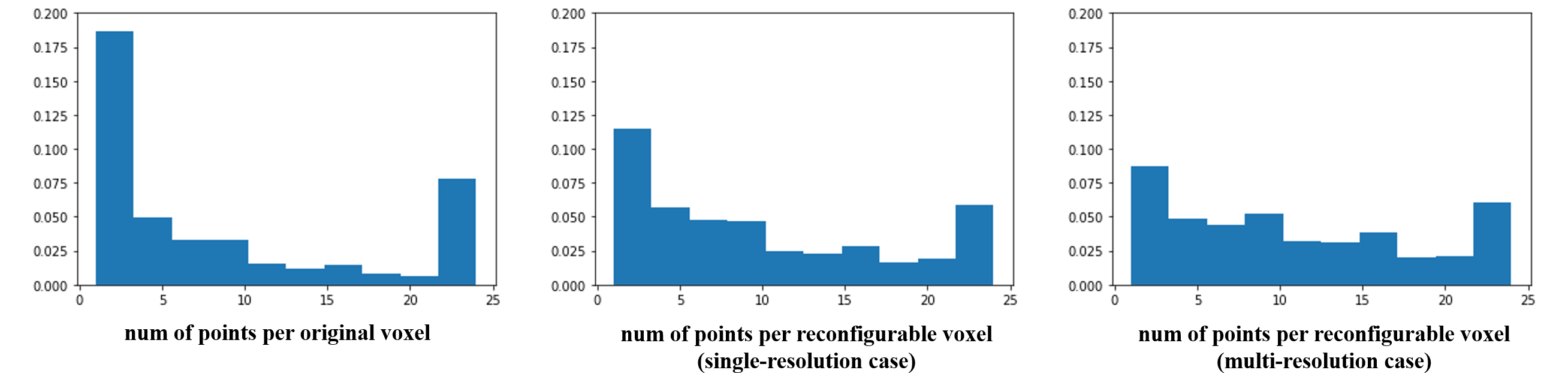}
\end{center}
    \vspace{-1ex}
    \caption{Change in distribution w.r.t number of points per voxel.}
\label{fig: dist_change}
\end{figure*}

Fig.~\ref{fig: dist_change} shows the change in distribution with respect to number of points per voxel.\footnote{Note that number of points per reconfigurable voxel refers to the average number of points in 5 voxels contained in the representation.} It reveals that the imbalance problem is alleviated and the reconfiguration in the multi-resolution case works even better than the single-resolution case. Specifally, voxels containing only one point are reduced and some of them are changed into voxels with decent number of points meanwhile.

\section{Implementation Details}\label{network_details}
\begin{figure}
\begin{center}
\vspace{-1ex}
\includegraphics[width=0.8\linewidth]{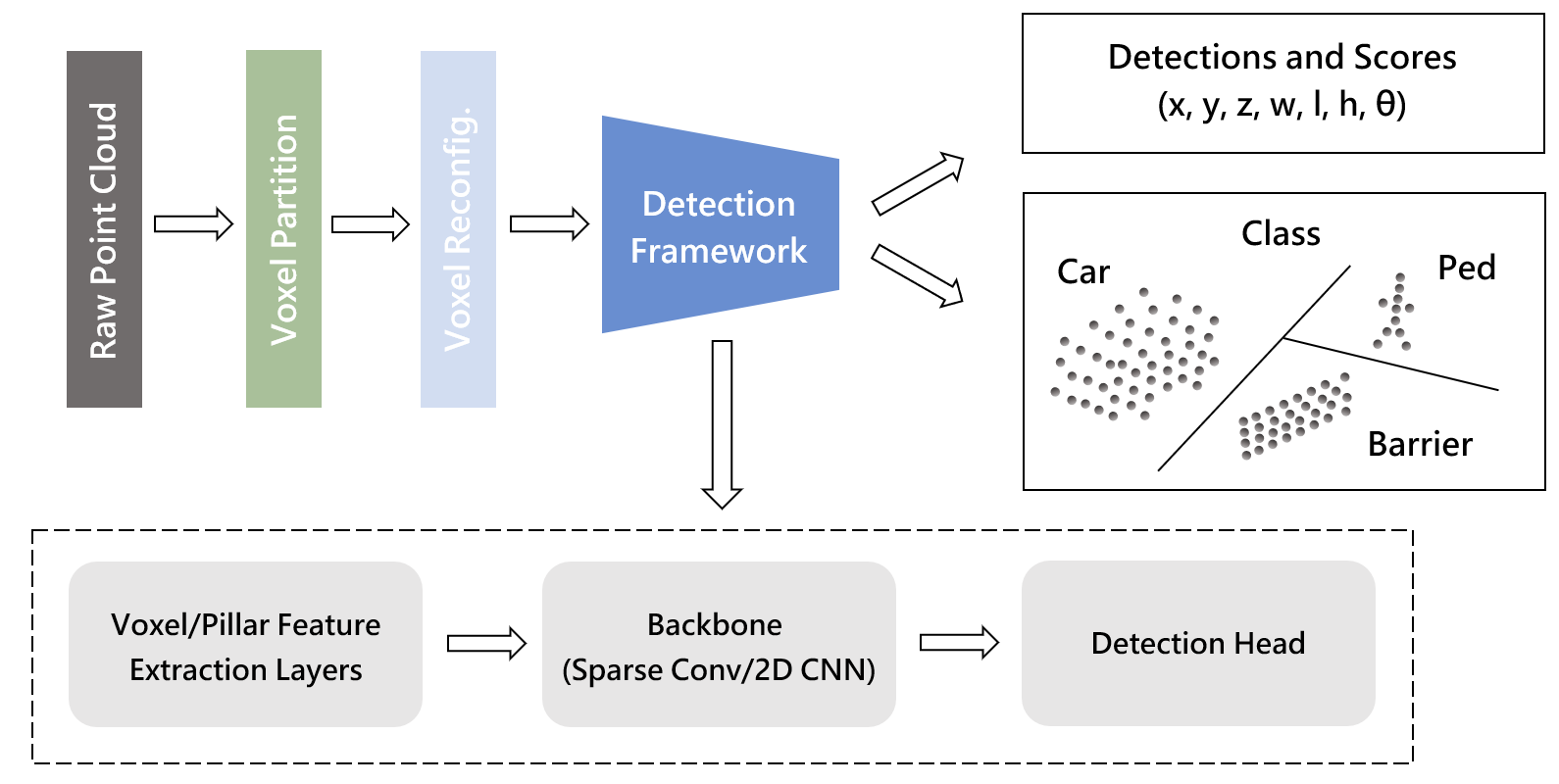}
\end{center}
    \vspace{-1ex}
    \caption{Overview of our framework.}
\label{fig: overview}
\vspace{-2.5ex}
\end{figure}

We basically follow two voxel-based frameworks, SECOND~\cite{SECOND} and PointPillars~\cite{PointPillars}, in the experiments. Different settings of these two frameworks are listed in Tab.~\ref{tab:diffset}. We follow the original method of PointPillars and SECOND in the setting of detection range and other details such as anchors and matching strategy except a few adjustments for network designs. Next in this section, we first present details of implementing random walk in different frameworks, and then elaborate the network designs in terms of three different modules shown in Fig.~\ref{fig: overview}.

Firstly, in terms of specific parameter setting of random walk, in order to ensure that the neighbor voxels will not go too far away due to random walk, we divide the number of points by 4 and round them up when we calculate Eqn. 1 and 2 in PointPillars, then the resulting steps range and the probability space of random walk are similar with the situation of SECOND. Details of Eqn. 1 and 2 can be referred to the main paper.

\begin{figure*}
\begin{center}
\includegraphics[width=1.0\linewidth]{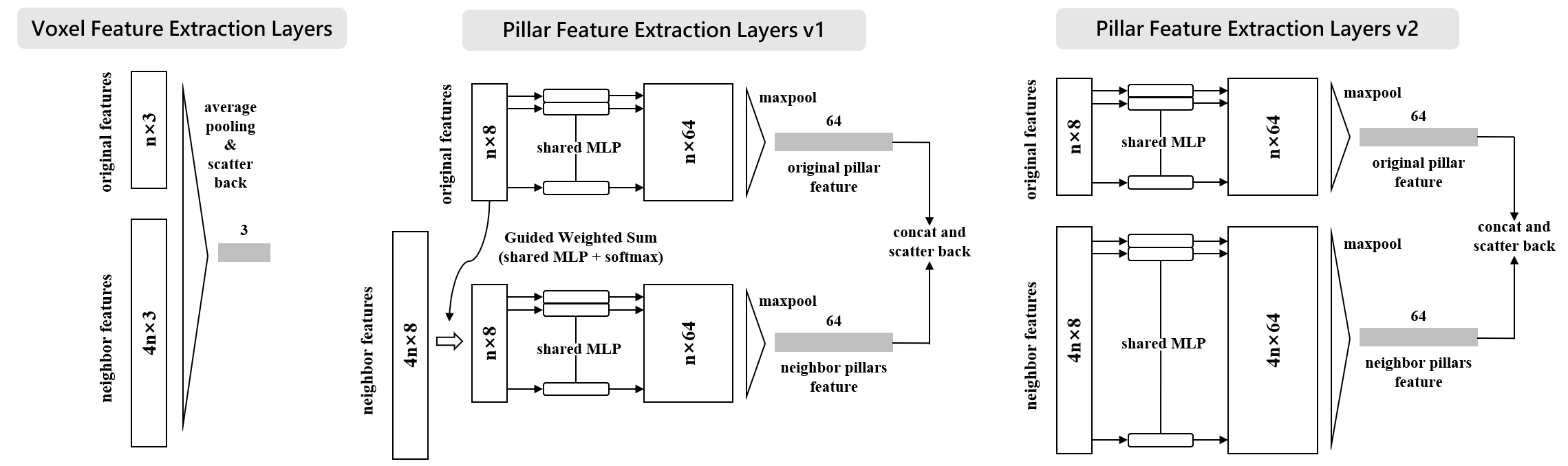}
\end{center}
    \vspace{-0.2cm}
    \caption{Voxel/Pillar Feature Extraction Layers.}
\label{fig: VFE_layers}
\end{figure*}

Then for network designs, we have stated basic ideas of the first module, feature extraction layers, in the paper (Sec. 3.3). To review, we briefly summarize it as follows. The voxel encoder in SECOND takes all the $5n$ points as input and processes their features by average pooling. In comparison, we tried two kinds of pillar feature extractors. Both of them encode features of original pillar and neighbor pillars respectively and then concatenate them along feature channels. The difference lies in the way of encoding the neighbor pillar features. The first version takes a weighted sum of these 4 neighbor pillar features at first while the second version directly encodes these $4n$ point features. As is shown in the results on the Lyft, the first version works better on the whole possibly because of the benefit from adaptive weights, and so it is adopted as our final implementation method. Specific parameters are shown in Fig.~\ref{fig: VFE_layers}. 

\begin{table*}
\vspace{-2ex}
\scriptsize
\begin{center}
\caption{Different settings in Reconfigurable PointPillars (Fast version) and SECOND experiments. To make experiments more comparable, PointPillars and SECOND in our experiments are both reproduced version, which share similar hyper parameters with the published ones while are improved on some details, thus have better performance in most cases. Our reconfigurable methods follow the same settings as well}
\label{tab:diffset}
\begin{tabular}[t]{c|c|c}
\hline
Method & PointPillars & SECOND \\
\hline\hline
Pillar/Voxel Input Features & $d$, $z$, $t$, $x_c$, $y_c$, $z_c$, $x_p$, $y_p$ & $d$, $z$, $r$ \\
\hline
Pillar/Voxel Resolution (m) & $0.25\times 0.25$ & $0.05\times0.05\times0.1$ \\
\hline
Max number of Pillars/Voxels & 25000 & 30000 \\
\hline
Max number of points per voxel & 25 & 4 \\
\hline
Convolutional Backbone & 2D CNN & 3D Sparse Convolution \\
\hline
Region Proposal Network & multi-group head & original RPN \\
\hline
\end{tabular}
\end{center}
\end{table*}

\begin{figure*}
\begin{center}
\includegraphics[width=1.0\linewidth]{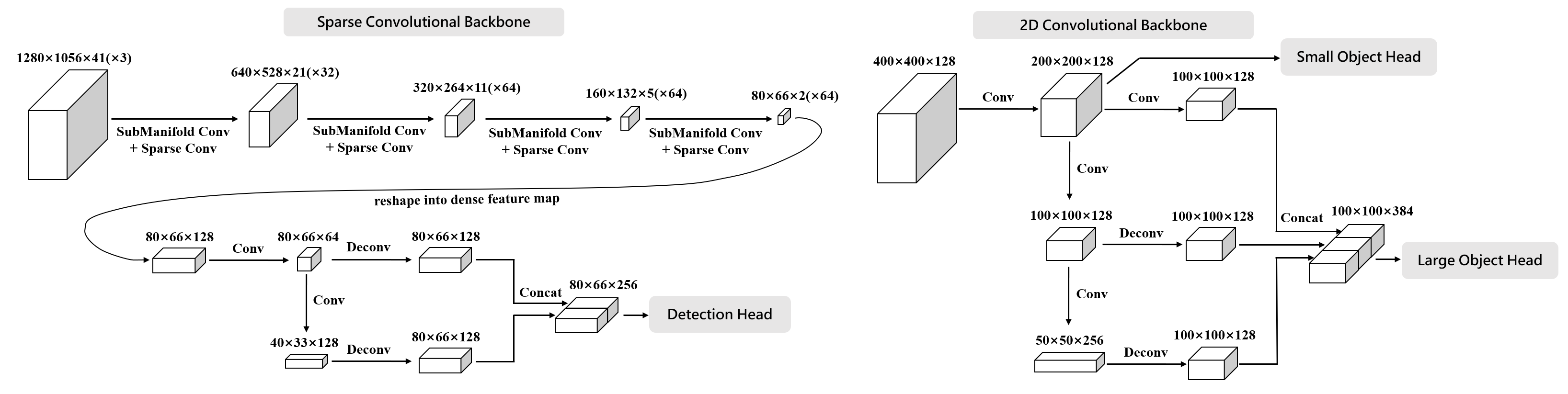}
\end{center}
    \caption{Backbone in SECOND and PointPillars.}
\label{fig: Backbone}
\end{figure*}

The second module, convolutional backbone is shown in Fig.~\ref{fig: Backbone}. In SECOND, the sparse convolutional backbone takes the 4D feature map as input and processes it by several submanifold convolution and sparse convolution layers. In Fig.~\ref{fig: Backbone}, the 4 downsampling steps consist of (2,1), (2,1), (3,1), (4,1) submanifold convolution and sparse convolution layers respectively. Then we utilize a top-down network to produce feature maps in two resolutions, perform upsampling and concatenation to derive the final feature map used in the subsequent detection head. In PointPillars, instead of preprocessing by sparse convolution layers, we just perform top-down convolution, upsampling and concatenation to get the input of large object head. In comparison, the shallow feature map is used to detect small objects. All the top-down convolutional layers contain 1 downsampling layer with stride = 2 and 5 layers with stride = 1 except the first convolution in PointPillars, which has 1 downsampling layer with stride = 2 and 3 layers with stride = 1.

The last module, detection head, just derives box map, class map and attribute map (if necessary) from the input feature map like SSD~\cite{SSD}. In PointPillars, the large object head is set the same as the head in SECOND while the small object head takes 3 extra convolutional layers to compress the feature map to 64 feature channels first. 

So far, we have introduced the details of network architectures. In conclusion, our implementation of SECOND is mainly different from that of PointPillars in terms of encoding methods, preprocessing in convolutional backbone and whether to design specific heads for different object categories. See the detailed structures and intermediate results in Fig.~\ref{fig: VFE_layers} and Fig.~\ref{fig: Backbone}.

\clearpage

\end{document}